\documentclass[letterpaper, 10 pt, conference]{ieeeconf}

\IEEEoverridecommandlockouts                            

\usepackage{cite}
\usepackage{amsmath,amssymb,amsfonts}
\usepackage{algorithmic}
\usepackage{graphicx}
\usepackage{comment}
\usepackage[dvipsnames]{xcolor}

\title{\LARGE \bf
Synthesized  Trust  Learning from Limited Human Feedback for Human-Load-Reduced Multi-Robot Deployments
}

\author{Yijiang Pang$^{}$, Chao Huang, Rui Liu$^{*}$
\thanks{* is the corresponding author {\tt\small ruiliu.robotics@gmail.com}.}
\thanks{$^{}$ Authors are with The Cognitive Robotics and AI Lab (CRAI), College of Aeronautics and Engineering, Kent State University, Kent, OH 44240, USA.}
}

\begin{document}

\maketitle
\thispagestyle{empty}
\pagestyle{empty}

\begin{abstract}
Human multi-robot system (MRS) collaboration is demonstrating potentials in wide application scenarios due to the integration of human cognitive skills and a robot team's powerful capability introduced by its multi-member structure. 
However, due to limited human cognitive capability, a human cannot simultaneously monitor multiple robots and identify the abnormal ones, largely limiting the efficiency of the human-MRS collaboration. There is an urgent need to proactively reduce unnecessary human engagements and further reduce human cognitive loads. 
Human trust in human MRS collaboration reveals human expectations on robot performance. Based on trust estimation, the work between a human and MRS will be reallocated that an MRS will self-monitor and only request human guidance in critical situations.
Inspired by that, a novel Synthesized Trust Learning (STL) method was developed to model human trust in the collaboration. STL explores two aspects of human trust (trust level and trust preference), meanwhile accelerates the convergence speed by integrating active learning to reduce human workload. 
To validate the effectiveness of the method, tasks "searching victims in the context of city rescue" were designed in an open-world simulation environment, and a user study with 10 volunteers was conducted to generate real human trust feedback.
The results showed that by maximally utilizing human feedback, the STL achieved higher accuracy in trust modeling with a few human feedback, effectively reducing human interventions needed for modeling an accurate trust, therefore reducing human cognitive load in the collaboration.
\end{abstract}

\section{INTRODUCTION}
Human multi-robot system (MRS) collaboration demonstrates great potentials in broad application scenarios benefiting from the integration of human cognitive skills and MRS's powerful capability introduced by its multi-members with diverse functions. With human guidance on robot motion correction, task reallocation, risk assessment, and status prediction, MRS performance is improved with more resiliency to real-world disturbances \cite{He2017survey, Doroodgar2010search, Hatanaka2015Passivity}.
In real-world applications, by monitoring the task process and robot performance, human involvements mainly manifested as giving out suggestions or directly manipulating the robots \cite{Kolling2013Human, liu2019trust}.
However, wide implementations of human-MRS collaboration are still limited by the human cognitive capability. First, uncertainty happens during the robots performing tasks, it is hard for a human to shift attention between multiple faulty factors and robots, especially when the task environment is complex or the number of robot requesting for human assistance is large.
Second, a long time of supervision on robot performance imposes heavy cognitive loads on human operators, making them tired and uncomfortable and therefore reducing the social acceptance of an MRS. There is an urgent need to make robots understand human expectations to proactively reduce unnecessary human engagements and further reduce human cognitive loads for effective collaboration.

\begin{figure}[t!]
  \centering
  \includegraphics [width=0.9\columnwidth ]{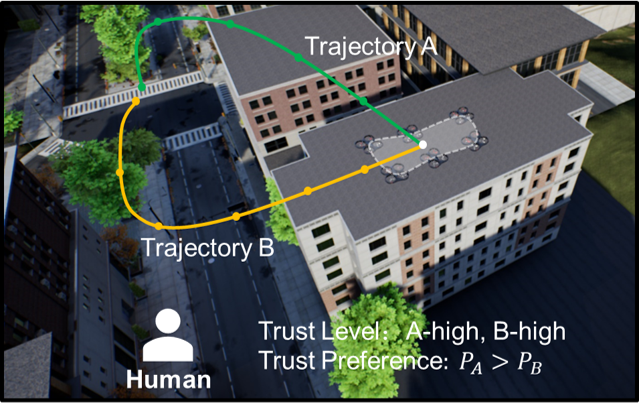}
  \caption{The illustration of human trust manifesting as trust level and trust preference. Human operator rates trajectory $A$ and trajectory $B$ with trust level "high" and shows preference with "$P_{A}>P_{B}$".}
  \label{figure1}
\end{figure}

Human trust in human MRS collaboration shows human expectation on robot performance \cite{Abbass2018Foundations, Billings2012interaction}. Investigating human trust is crucial for developing effective cooperation between humans and an MRS, such as helping robots to understand human expectations, actively identifying unsatisfying robot behaviors, and selectively reporting critical issues to request human corrections.
Based on the trust estimation, the work between a human and an MRS will be reallocated that an MRS will self-monitor its performance, reducing unnecessary human interventions and only requesting human guidance in critical situations. Therefore to solve the human supervision and cognitive load challenges, this paper developed an intelligent trust model to estimate human trust to help an MRS to improve the collaboration with the avoidance of unnecessary human assistance.

In real-world situations, human trust manifests as two different types, trust level and trust preference. As shown in Fig \ref{figure1}, the trust level is a human operator's assessment for robot mission performance, which is represented as discrete values usually, adapting to human's discrete representations \cite{Dietrich2003Discrete}. 
Trust preference is a human operator's comparison between two different performances of robots performing tasks, which reveals human criteria of adjusting trust based on robot behaviors. Therefore the trust preference is more informative in understanding human expectations comparing with the absolute trust level.

In this paper, a novel \textbf{\textit{Synthesized Trust Learning}} (STL) method was developed to model human trust in the collaboration, which includes two aspects of human trust, trust level and trust preference, meanwhile, accelerates the convergence speed by integrating active learning to reduce human workload.
This paper mainly has three contributions:
\begin{enumerate}
\item A novel trust modeling method, \textbf{\textit{Synthesized Trust Learning}}, is developed to maximally explore heterogeneous human feedback to model human trust. \textbf{\textit{STL}} enables a human-robot cooperative system to quickly model the human operator's expectation on the performance of robots and further assist human operators in guiding robots.    

\item An accelerating learning manner is designed through deep integration between active learning framework with the proposed method and an open-world simulation environment. 

\item A novel trust-aware partnership is designed to facilitate human MRS collaboration based on trust estimation. The trust model integrating human expectation actively assists a human with monitoring robot performance, reducing the burden of human operator.
\end{enumerate}

\section{RELATED WORK}
Prior researches investigated the computational trust models \cite{Hussein2020Towards}.
In \cite{Mahani2020Bayesian}, the updating process of degrees of trust between the human operator and multiple robots was considered as a factorial hidden Markov model (HMM). Robot performance that was quantified with a linear model parameterized with performance features, human-robot interaction data and subjective assessments from the human operator were used to update the degree of trust in robot.
\cite{Soh2019Trust} proposed Bayesian and recurrent neural models to predict self-reported human trust that changed with observations of robot performance.
\cite{Nam2019Models} developed a trust model to quantify the trust level between human operator and robot swarm, which considered the trust level as a weighted linear combination of performance features.
However, the above-mentioned researches ignored the trust information contained in the trust preference, which limits their model performance in the situations where require a comparison between similar task performances. In this paper, \textbf{\textit{STL}} maximally explores heterogeneous human feedback including both trust rating and comparison to model human trust.

Researches have been done to learn from heterogeneous human feedback.
\cite{Sadigh2017Active} developed a general active preference-based method to extract information from human preference between two trajectories. The ratings of two trajectories were modeled as selection probability, then active learning and sampling methods were used to calculate weights for the feature.
\cite{Palan2019Learning} assumed reward function was a linear combination of features and explored data of human demonstrations and preferences. Their method combined a numerical optimization method that is easy and fast but usually subject to a local optimum with sampling methods to get the weights of features.
Different from this paper, the above approaches relied on and applied to linear regression models or polynomial regression models, which are hard to design with data having complex mapping relationship especially in scenarios of modeling human cognition. In this paper, \textbf{\textit{STL}}, which is based on deep learning, is more adaptive in complex concept learning.

Different from our paper, which focuses on continual learning with heterogeneous data, 
\cite{Xue2019Deepfusion} used different blocks of convolutional neural network to learn from different sensors to uniform the heterogeneous inputs.
\cite{Aljundi2018Memory, Mallya2018Packnet, Kirkpatrick2017Overcoming} achieved learning without forgetting by penalizing major changes in the parameters that were important for the previous tasks. In this paper, the \textbf{\textit{STL}} uses a shared structure to deal with heterogeneous data and achieves continual learning by maintaining the previous model performance on new tasks.

Our previous work explored the potentials of trust in human-robot cooperative system \cite{liu2019trust, liu2019trustRepair}. This paper further developed a more advanced trust model to actively assist human operators in the collaboration with a unique perspective of reducing human cognitive loads in collaboration.

\begin{figure*}[t!]
  \centering
  \includegraphics [width=0.9\linewidth ]{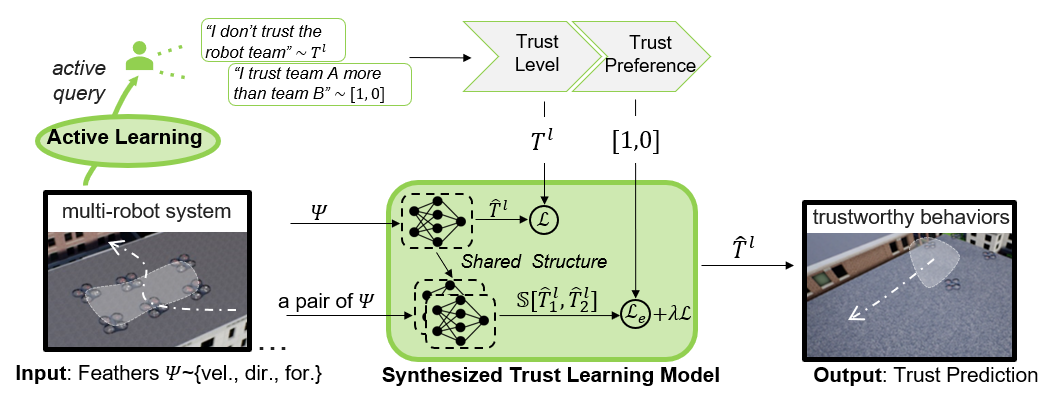}
  \caption{The architecture of the \textbf{\textit{Synthesized Trust Learning}} method. Human operator provide trust feedback by monitoring MRS performing tasks, then the heterogeneous data, \{features of a single trajectory and corresponding trust level, and features of a pair of trajectories and corresponding trust preference\}, are both used to update the trust model sequentially.}
  \label{figure2}
\end{figure*}

\section{PRELIMINARIES}
This section introduces the notations used throughout the paper and gives a brief overview of multi-robot systems, human trust feedback, and supervised learning setup of this paper.

\textbf{Multi-Robot Systems:} Consider a robot team of $n$ robots with status ${X}_{i}$, where ${X}_i = ({V}_{i},{P}_{i},{O}_{i})$. For robot $i$, ${V}_{i}\in \mathbb{R}^3$ denotes velocity, ${P}_{i}\in \mathbb{R}^3$ denotes position, and ${O}_{i}\in \mathbb{R}^3$ denotes orientation.
A task trajectory $\xi \in \Xi$ is a finite sequence of robot states, i.e., $\xi = (X)_{t=0}^{T}$, where $T$ is the end of the task. And the features of a trajectory is denoted as $\psi$, which is calculated by a feature extraction function $\psi=\phi(\xi)=\phi((X)_{t=0}^{T})$.

\textbf{Human Trust Feedback:} Human operator monitors task process and gives trust feedback according to the task performance of trajectories. ${T}^{level}$ denotes the trust level human rated for a trajectory, which is discrete and varies within $[-1, 1]$. For example, ${T}^{level}$ is one of ${T}^{demarc}=(-1, -0.5, 0, 0.5, 1)$ for five categories of trust level. Human operator can also provide trust preference to a pair of trajectories $[I, I^{\prime}]$. If the first trajectory is preferred,  $[I,I^{\prime}]=[1,0]$, or $[I,I^{\prime}]=[0,1]$ if the second trajectory is preferred.

\textbf{Supervised Learning Setup:} Consider a deep neural network based model represented by a parametrized function $f_{\theta}$ with parameters $\theta$. When adapting to a new task $\mathcal{T}_{a}$ with a corresponding dataset, $\mathbb{D}=\{(x_{i},y_{i})_{i=1}^{k}\}$ with $k$ example pairs, the model’s parameters $\theta$ become $\theta_a$. 
\begin{equation}
    \theta_a = \theta - \alpha\nabla_{\theta}\mathcal{L}_{\mathcal{T}_{a}}(f_{\theta})
\end{equation}
In this paper, $\mathcal{T}=\{\mathcal{T}_{a}, \mathcal{T}_{b}\}$, and $\mathcal{T}$ are learned sequentially, which turns the learning process to a continual learning problem. 
\begin{equation}
    \theta_{b} = \theta_{a} - \beta\nabla_{\theta_{a}}\mathcal{L}_{\mathcal{T}_{b}}(f_{\theta_{a}})
\end{equation}
where the $\alpha, \beta$ is the learning rate.
Specifically, $\mathcal{T}_{a}$ and $\mathcal{T}_{b}$ have heterogeneous data. 
$\mathcal{T}_{a}$ consists of the input $\psi$ and the corresponding trust level ${T}^{level}$ as the label, and is with a corresponding dataset,
$\mathbb{D}_{a}=\{(x_{i},y_{i})_{i=1}^{m}\}=\{(\psi_{i},{T}^{level}_{i})_{i=1}^{m}\}$.
$\mathcal{T}_{b}$ consists of the input a pair of $\psi$ and the corresponding trust preference $(I, I^{\prime})$ as the label, and is with a corresponding dataset,
$\mathbb{D}_{b}=\{(x_{i},y_{i})_{i=1}^{n}\}=\{([\psi_{i}^{p},\psi_{i}^{q}],[I_{i}, I_{i}^{\prime}])_{i=1}^{n}\}$.

\section{SYNTHESIZED TRUST LEARNING}
This section describes the \textbf{\textit{Synthesized Trust Learning}} method, which maximally explores heterogeneous human feedback and accelerates the convergence speed by integrating active learning mechanism to model human trust in a human-robot cooperative system.
The architecture of the \textbf{\textit{STL}} is shown in Fig \ref{figure2}.
\subsection{Continual Learning with Heterogeneous Data}
The \textbf{\textit{STL}} is a two-step continual learning method. 
Firstly, the model learns from task $\mathcal{T}_{a}$ with dataset $\mathbb{D}_{a}=\{(x_{i},y_{i})_{i=1}^{m}\}=\{(\psi_{i},{T}^{level}_{i})_{i=1}^{m}\}$, and the model's parameters $\theta$ become $\theta_{a}$. Secondly, the model learns from task $\mathcal{T}_{b}$ with dataset $\mathbb{D}_{b}=\{(x_{i},y_{i})_{i=1}^{n}\}=\{([\psi_{i}^{p},\psi_{i}^{q}],[I_{i}, I_{i}^{\prime}])_{i=1}^{n}\}$, where $[\psi^{p},\psi^{q}]$ denotes a pair of trajectories, and the model's parameters $\theta_{a}$ become $\theta_{b}$.

\textbf{Learning Task $\mathcal{T}_{a}$:}
The goal of this step is to get a trust level prediction model. 
The object function that we minimize in $\mathcal{T}_{a}$ is
\begin{equation}
\begin{aligned}
    \mathcal{L}_{\mathcal{T}_{a}}(f_{\theta})=\sum_{\psi_{i},{T}^{level}_{i}\sim \mathcal{T}_{a}}\mathcal{L}_{2}(f_{\theta}(\psi_{i}), {T}^{level}_{i})
\end{aligned} 
\end{equation}
where the output $f_{\theta}(\psi_{i})=\hat{T}^{level}$ is continuous and varies within $[-1,1]$, and the $\forall{T}^{demarc}$ that is closest to $\hat{T}^{level}$ will be considered as the trust level for the input trajectory. However, this model $f_{\theta_{a}}$ ignores the trust information between different trajectories that are on the same trust level.

\textbf{Learning Task $\mathcal{T}_{b}$:}
The goal of this step is considered as getting a trust model, which can predict trust level for a single trajectory and distinguish the trust preference for a pair of trajectories even in the same trust level.
This paper assumes that the human's trust preference is with respect to the true trust level of each task trajectory. As per this model, the trust preference that the human gives to a pair of trajectories is calculated by $softmax(f_{\theta}(\psi^{p}), f_{\theta}(\psi^{q}))$.

The loss function \cite{li2017Learning} that we minimize in $\mathcal{T}_{b}$ is
\begin{equation}
\begin{aligned}
    &\mathcal{L}_{\mathcal{T}_{b}}(f_{\theta})=\\
    &\sum_{[\psi_{i}^{p},\psi_{i}^{q}],[I_{i}, I_{i}^{\prime}]\sim \mathcal{T}_{b}}\mathcal{L}_{e}(softmax(f_{\theta}(\psi_{i}^{p}), f_{\theta}(\psi_{i}^{q})), [I_{i}, I_{i}^{\prime}])\\
    &+\lambda \mathcal{L}_{2}(f_{\theta}(\psi_{i}^{p}),f_{\theta_{a}}(\psi_{i}^{p}))+\lambda \mathcal{L}_{2}(f_{\theta}(\psi_{i}^{q}),f_{\theta_{a}}(\psi_{i}^{q}))
 \end{aligned} 
\end{equation}
where $\mathcal{L}_{e}$ is cross entropy loss function.
$\mathcal{L}_{e}(softmax(f_{\theta}(x^{(i)})), y^{(i)})$ makes the output be fit for the ground truth of trust preference;
$\lambda \mathcal{L}_{2}(f_{\theta}(x^{(i)}),f_{\theta_{a}}(x^{(i)}))$ is the continual learning term, which compensates the output to be fit for the output of original trust level of $\mathcal{T}_{a}$;
$\lambda$ is a loss balance weight. 

\subsection{Active Learning}
In order to speed up convergence further reduce human workload, this paper synthesizes experiments with active learning mechanism for $\mathcal{T}_{a}$. 
The constraint combines the least confidence strategy with maximum difference strategy, which means the model before applying active learning has the least confidence in its most likely label for the generated trajectories, besides, the generated data should have the maximum difference with the existed data.
\begin{equation}
\begin{aligned}
    &\underset{\psi}{\text{minimize: }}|f_{\theta}(\psi)-\forall T^{demarc}|+\lambda S_{C}(\psi, \forall\Psi^{d})\\
    &\text{subject to: }\psi \in \Psi
\end{aligned} 
\end{equation}
where $S_{C}$ is cosine similarity, $\Psi^{d}$ is the set of existed training data, and $\Psi$ is set of feasible data.

\section{EVALUATION}
In order to validate the effectiveness of \textbf{\textit{STL}} in modeling human trust in human-robot cooperative system, a task scenario, "multiple robots search victims in the context of city rescue", was designed in an open-world simulation environment. 
And a pioneer user study was conducted to collect human trust feedback of task trajectories for model training and validation.
The following aspects were validated. 
(i)the effectiveness of \textbf{\textit{STL}} in reducing the human workload of labeling;
(ii)the effectiveness of \textbf{\textit{STL}} in improving model performance with heterogeneous data learning.
\subsection{Experiment}

\begin{figure}[t!]
  \centering
  \includegraphics [width=1\columnwidth ]{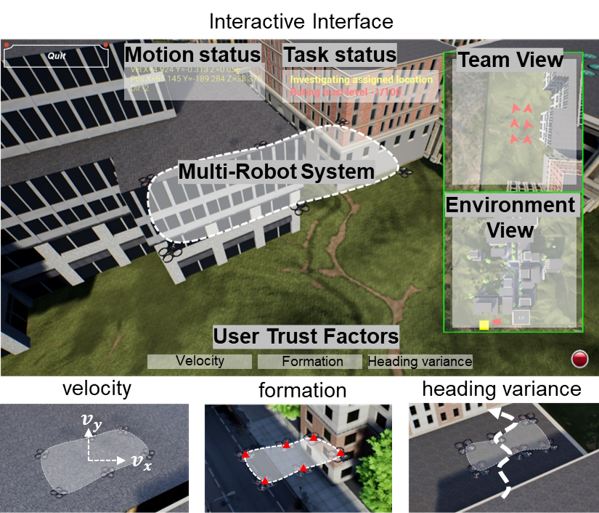}
  \caption{The illustration of the experiment setting, including an interactive interface and three typical robot behaviors that influence human trust.}
  \label{env}
\end{figure}

\textbf{Environment Setting}
This paper envisioned a real-world human-robot cooperative system where human operator was able to monitor robot motion status, task status, and had a higher level view of observing the surrounding environment and overall environment supported by panoramic images and positioning system.

There were three main sub-parts in the experiment, \textit{simulation environment}, \textit{interactive interface}, and \textit{multi-robot system}, as shown in Fig \ref{env}.
The open-world simulation environment was developed based on open-source platforms AirSim \cite{Shah2018airsim} and Unreal Engine \cite{epic2019unreal}, and the map described an urban city environment. 
Human operator interacted with the environment through a customized interactive interface, shown in the "Interactive Interface" of Fig \ref{env}. The main view of the interface displayed the multi-robot system; some customized widgets displayed the motion status, task status, team view, and overall environment view to let the human operator learn the task process and performance better. 
The multi-robot system consisted of six Unmanned Aerial Vehicles (UAVs). In order to support active learning, this paper developed independent application program interfaces to control three features of the robot team, velocity, formation, and heading direction variance.

In each task, one or two target locations were randomly generated to simulate the victim locations, and another parameters, $\psi_{c}$, were randomly generated in the preset range to control the above-mentioned three features of the robot team. The robot team was initialized to a same starting location then flew to the assigned target locations automatically supported by an integrated path planning algorithm. 
During a task, the task trajectory, $\xi = (X)_{t=0}^{T}$, would be recorded to a local file, and in this paper, the feature extraction function $\phi(\xi)$ extracted average velocity, formation, and heading direction variance from the local file as the real trajectory features, $\psi$.
Meanwhile, human operator observed the task process and task performance through the interactive interface.

\textbf{Human User Study}
A human user study consisting of 10 participants was conducted. 
The user study comprised two main parts, a tutorial and experimental surveys collected after viewing all randomly generated trajectories. The tutorial included two video examples of two sample tasks belonging to $\{\mathcal{T}_{a}, \mathcal{T}_{b}\}$ respectively. The experimental survey had a series of sequential task trajectories and pairs of task trajectories. Participants were asked to observe the trajectories and then provide trust feedback. 
\begin{figure}[t]
  \centering
  \includegraphics [width=1\columnwidth ]{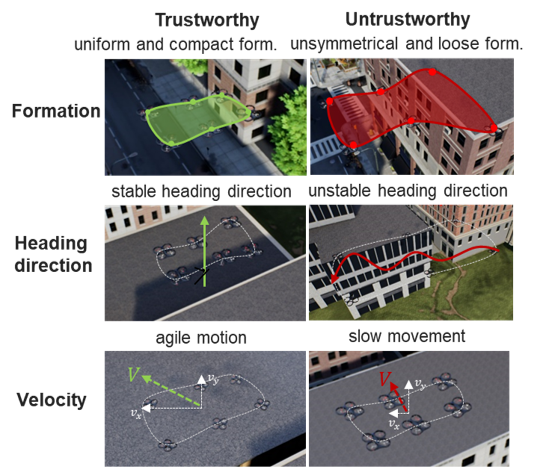}
  \caption{The illustration of the sample robot behaviors with predicted trustworthiness.}
  \label{behaviors}
\end{figure}
\subsection{Result analysis}
Overall, the illustration of the sample robot behaviors with predicted trustworthiness was summarized in Fig. \ref{behaviors}, showing the trust model's potential in assisting human to monitor task performance, actively reducing the burden of human operator.

\begin{figure}[t!]
  \centering
  \includegraphics [width=1\columnwidth ]{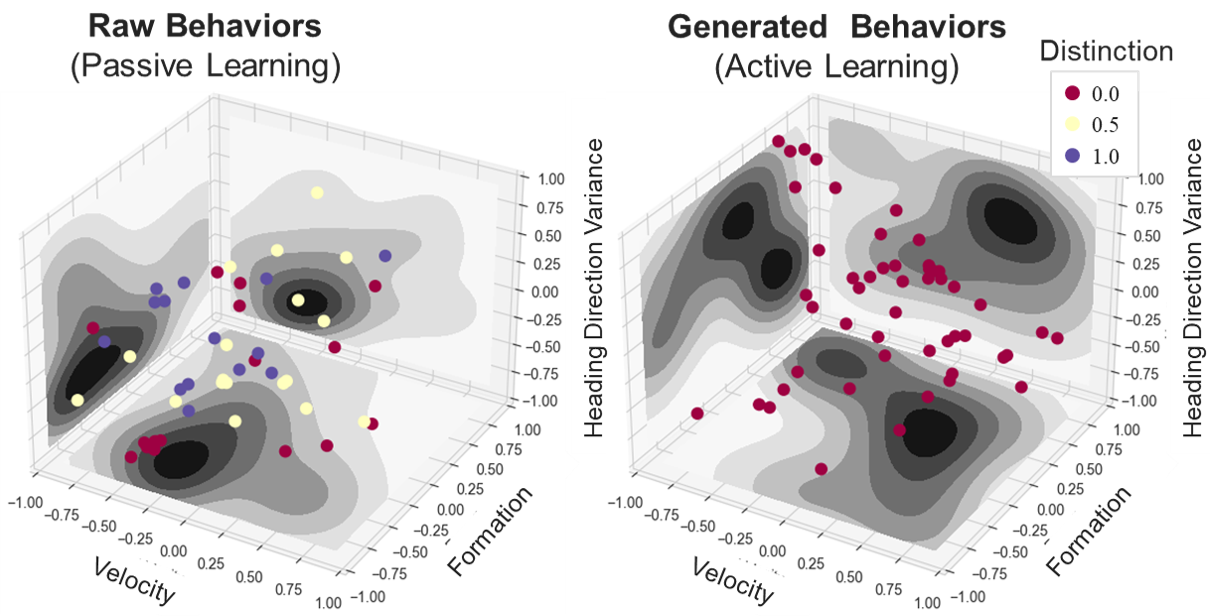}
  \caption{The comparison between raw task trajectory features and generated task trajectory features. Each dot represents a trajectory feature, and the color of a dot represents the degree of distinction of the dot for the current model. The density distribution of all dots is projected onto each axis.}
  \label{generated}
\end{figure}

\textbf{Validation of Active Learning Performance.}
In order to visualize the difference between raw trajectory features and the trajectory features generated by the active learning, 50 copies of raw trajectory features were selected to compare with the generated features, as shown in Fig. \ref{generated}.
The degree of distinction, which is calculated by $min|f_{\theta}(\psi)-\forall T^{demarc}|$ and scaled to three levels $[1.0, 0.5, 0.0]$, and the difference of data distribution, which is measured by two-dimensional Kolmogorov–Smirnov test \cite{Fasano1987kstest}, were used to measure the effectiveness of active learning, which are the reflections of least confidence strategy and maximum difference strategy respectively.
(i) data distribution.
The distribution of raw trajectory features and the distribution of uniform features in three axis were significantly different $(p_{xy}<0.05, p_{xz}<0.05, p_{yz}<0.05)$ in the data set of raw trajectory features;
The distribution of the combined features and the distribution of uniform features in three axis were similar $(p_{xy}>0.1, p_{xz}>0.1, p_{yz}>0.1)$ in the data set of raw trajectory features plus generated trajectory features.
Therefore, the active learning method can enrich the data diversity by compensating the raw data with maximum difference strategy, which can increase model convergence speed and further reduce human workload by providing model with unfamiliar but task-related knowledge.
(ii) degree of distinction.
The number of trajectories with different distinction levels were 14, 18, and 18, respectively in the data set of raw trajectory features. While in the data set of trajectories generated with least confidence strategy, the current sub-optimal model had a relatively small distinction level for all the trajectories, showing that the generated trajectories contain knowledge that is unfamiliar for the current sub-optimal model. Therefore, the active learning mechanism can speed up the training process by actively requesting human assistance for learning unfamiliar but task-related knowledge, which therefore reduce the total number of samples needed and relief human cognitive workload.

\begin{figure}[t!]
  \centering
  \includegraphics [width=1\columnwidth ]{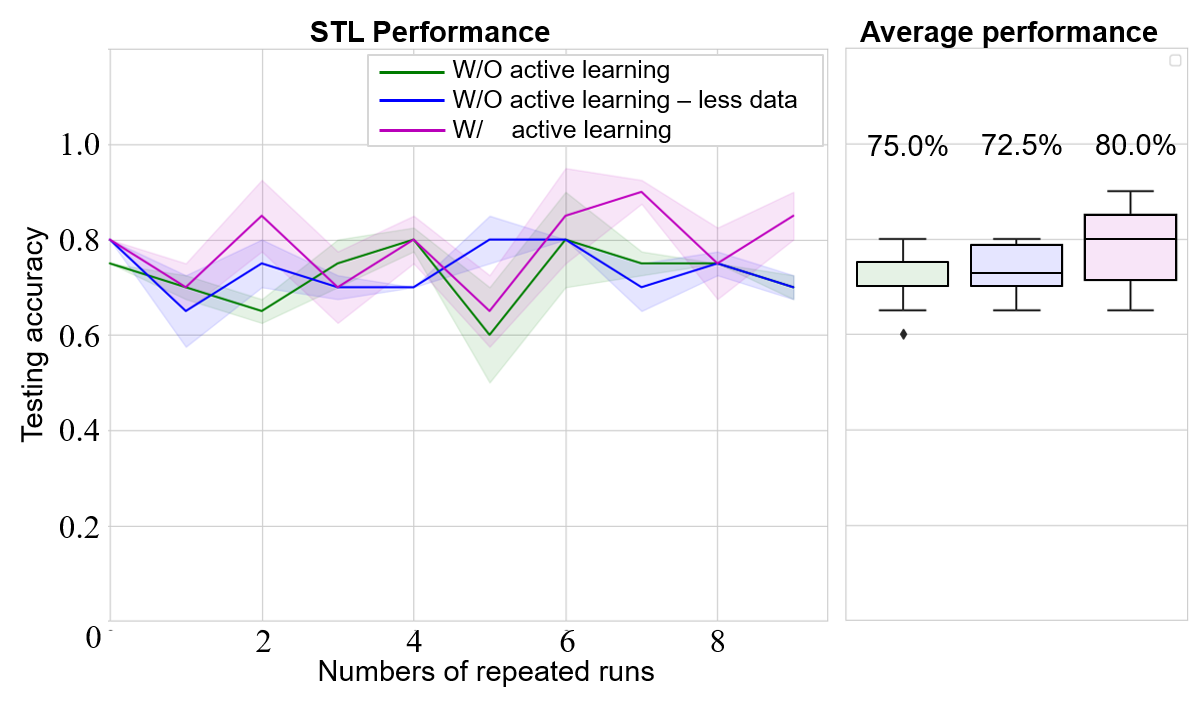}
  \caption{The comparison of testing accuracy in prediction of trust level between the model using active learning mechanism and without using active learning mechanism. It shows the model using the proposed active learning mechanism had a better performance in the prediction of trust level than the model without using active learning mechanism when training with the same number of training samples.}
  \label{active}
\end{figure}

\textbf{Validation of Human Workload Reduction.}
In order to validate the effectiveness of \textbf{\textit{STL}} in reducing human workload of labeling, the model used three different settings of training data to train the model, as shown in Fig. \ref{active}.
"W/O active learning -less data" was trained with 40 training samples sampled from $\mathbb{D}_{a}$;
"W/O active learning" was trained with 60 training samples sampled from $\mathbb{D}_{a}$;
"W/ active learning" was trained with 40 training samples sampled from $\mathbb{D}_{a}$, and another 20 training samples generated by active learning method then labeled individually.
Each setting was run ten times to mitigate the uncertainty introduced by random data sampling and random initialization of neural network parameters. The setting, "W/ active learning", reached the highest average test accuracy, 80$\%$, as shown on the right of Fig. \ref{active}. 
The result indicated that the model using the proposed active learning mechanism had a better performance in the prediction of trust level than the model without using active learning mechanism when training with the same number of training samples (same amount of human workload), testing accuracy increased 5$\%$. 

\begin{figure}[t!]
  \centering
  \includegraphics [width=1\columnwidth ]{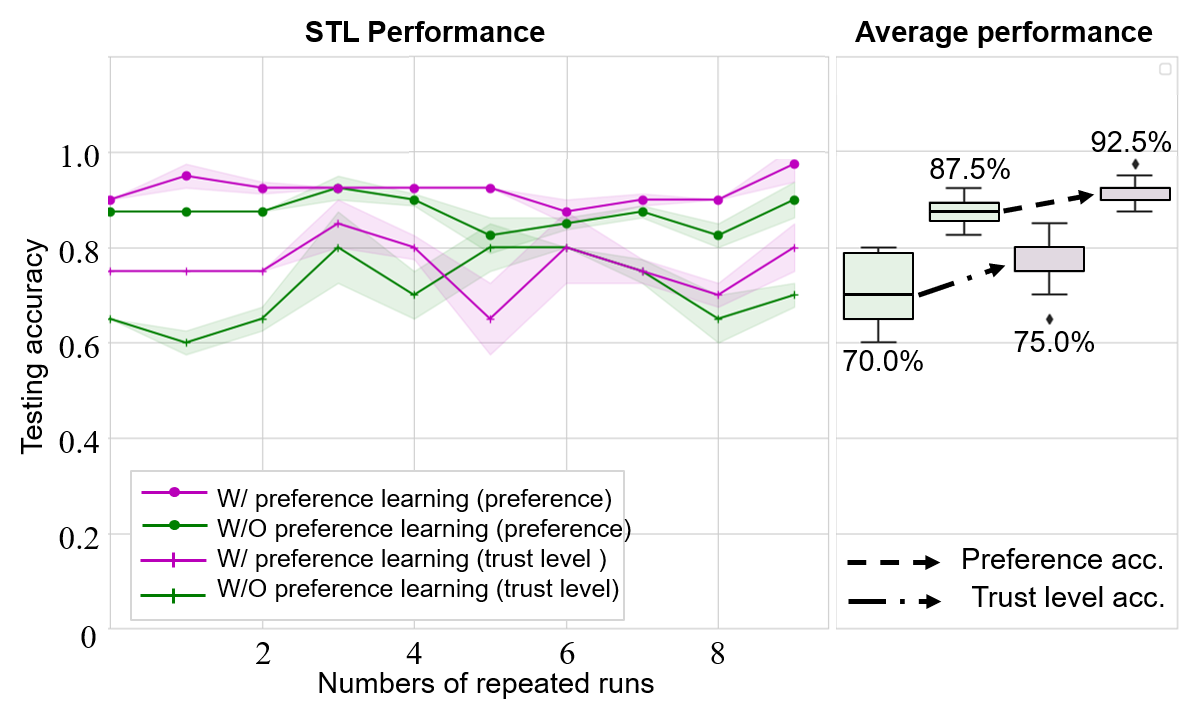}
  \caption{The comparison of testing accuracy in prediction of trust level and trust preference between the model using heterogeneous data learning and without using heterogeneous data learning. It shows the model using the proposed heterogeneous data learning had a better performance in the prediction of trust level and trust preference than the model without using active learning mechanism.}
  \label{preference}
\end{figure}

\textbf{Validation of Heterogeneous Data Learning}
Trust preference is more informative compared with trust level, which reveals human criteria of adjusting trust by providing a comparison between two trajectories with the same trust level.
By combining the two aspects of human trust (trust level and trust preference), the real human trust could be revealed more comprehensively, therefore, facilitating the alignment between the real human trust and the learned trust model. 
In order to validate the effectiveness of \textbf{\textit{STL}} in improving model performance with heterogeneous data learning, the model used two different settings of training data to train the model.
"W/O preference learning" was trained with 60 training samples sampled from $\mathbb{D}_{a}$;
"W/ preference learning" was trained with 60 training samples sampled from $\mathbb{D}_{a}$, and another 40 training samples sampled from $\mathbb{D}_{b}$.
As shown in Fig. \ref{preference}, each setting was run ten times to mitigate the uncertainty introduced by random data sampling and random initialization of neural network parameters. 
The setting, "W/ preference learning", reached 92.5$\%$ testing accuracy in prediction of trust preference and 75$\%$ testing accuracy in prediction of trust level; The setting, "W/O preference learning", reached 87.5$\%$ testing accuracy in prediction of trust preference and 70.0$\%$ testing accuracy in prediction of trust level.
The result indicated that the model using the proposed heterogeneous data learning method learned the trust preference information and further improved model performance in prediction of trust level.

\section{CONCLUSION AND FUTURE WORK}
This paper developed a \textbf{\textit{Synthesized Trust Learning}} method to model human trust in human-robot cooperative system, which maximally explored two aspects of human trust, trust level and trust preference.
To validate the effectiveness of \textbf{\textit{STL}}, this paper envisioned a real-world human-robot cooperative system and developed a simulation environment, interactive interface, and programmable multi-robot system to support the task scenario, "Multiple robots search victims in the context of city rescue". A user study with 10 volunteers was conducted to collect trust feedback for training and testing the trust model.
The effectiveness of \textbf{\textit{STL}} in reducing human workload was validated by a higher accuracy in the prediction of trust level than the baseline method when using the same amount of human workload.
The effectiveness of \textbf{\textit{STL}} in improving model performance with heterogeneous data learning was validated by improved model performance in the prediction of trust level and trust preference.
In the future, more patterns of human trust will be investigated to expand the ability of the trust model; then the feature extraction function, $\phi(\xi)$, can be improved to better extract trajectory features.

\end{document}